\documentclass[final,3p,times]{elsarticle}
\usepackage{longtable}
\usepackage{hyperref}
\usepackage{xcolor,soul, booktabs}

\usepackage{multirow,caption,array}
\newcolumntype{P}[1]{>{\centering\arraybackslash}p{#1}}
\usepackage{supertabular}
\usepackage{amsmath}

\usepackage{graphicx,subcaption}

\journal{Journal of \LaTeX\ Templates}

\begin{document}

\begin{frontmatter}

\title{Physico-chemical properties extraction from the fluorescence spectrum with 1D-convolutional neural networks: application to olive oil}

\author[zhaw,toelt]{Francesca Venturini\corref{mycorrespondingauthor}}
\ead{vent@zhaw.ch}
\author[polito]{Michela Sperti}
\author[toelt]{Umberto Michelucci}
\author[toelt,lju]{Arnaud Gucciardi}
\author[granada]{Vanessa M. Martos}
\author[polito]{Marco A. Deriu}

\cortext[mycorrespondingauthor]{Corresponding author}

\address[zhaw]{Institute of Applied Mathematics and Physics, Zurich University of Applied Sciences, Technikumstrasse 9, 8401 Winterthur, Switzerland}
\address[polito]{PolitoBIOMed Lab, Department of Mechanical and Aerospace Engineering, Politecnico di Torino,
Turin, Italy}
\address[toelt]{TOELT llc, Machine Learning Research and Development, Birchlenstr. 25, 8600 D\"ubendorf, Switzerland}
\address[lju]{Artificial Intelligence Laboratory, University of Ljubljana, Ljubljana, Slovenia}
\address[granada]{Department of Plant Physiology, Faculty of Sciences, Biotechnology Institute, Campus Fuentenueva s/n, 18071 University of Granada, Spain}

\begin{abstract} 
One of the main challenge for olive oil producers is the ability to asses and control oil quality regularly during the production cycle. For this purpose, several parameters need to be determined, such as the acidity, UV absorption coefficients and the ethyl esters content. To achieve this, samples must be sent to an approved laboratory for chemical analysis. This approach is expensive, required costly equipment and scientific training and thus cannot be performed very frequently, making regular and low-cost quality control of olive oil a hard to solve challenge.
This work describes a new approach based on fluorescence spectroscopy and artificial intelligence (namely, 1-D convolutional neural networks) to predict the five chemical quality indicators of olive oil (acidity, peroxide value, UV spectroscopic parameters $K_{270}$ and $K_{232}$, and ethyl esters) from one single fluorescence spectrum obtained with a very fast (less than one second) measurement from a low-cost fluorescence sensor. The measurement does not require any sample preparation or scientific training and can be easily performed even by completely untrained workers. The results indicate that the proposed approach gives exceptional results for  quality determination through  extraction of the relevant physico-chemical parameters. This would make the continuous quality control of olive oil during and after the entire production cycle a reality, even for small producers.

\end{abstract}

\begin{keyword}
artificial intelligence\sep neural networks\sep machine learning\sep fluorescence spectroscopy\sep optical sensor\sep olive oil \sep quality control
\MSC[2010] 00-01\sep  99-00
\end{keyword}

\end{frontmatter}



\section{Introduction}

Determining the quality of olive oil is an expensive and complex procedure that requires a chemical analysis by accredited laboratories and an organoleptic evaluation by accredited testing panels. For producers it is thus impossible to determine olive oil quality  effectively and frequently enough during the production process. This is important as olive oil chemical composition changes dramatically with time \cite{GOMEZCOCA2016378} due to, for example, storage and temperature conditions.
Chemical analysis are complex, time consuming, and require expensive equipment and scientific training that are not available to producers.  The chemical parameters, the procedures for their determination (methods ranging from titration to gas chromatography), as well as the limiting value for each olive oil quality class are specified in the European regulation \cite{regulation1991commission} and amendment \cite{european2013commission}. These regulations provide a decision tree for the verification if an olive oil class is consistent with the declared quality. This work describes a new technology based on a low-cost fluorescence sensor \cite{venturini2021exploration} and specificially designed one-dimensional convolutional neural networks \cite{Michelucci2017} that makes the chemical analysis low-cost, possible on-site and without the need of having scientifical training or expensive equipment. All the  chemical parameters necessary for the determination of the olive oil quality (Acidity, Peroxide Value, $K_{270}$, $K_{232}$ and Ethly esters) can be determined with one single measurement that is done on one unprepared and undiluted oil sample in less than one second. Thus making continuous olive oil quality control a reality for even small producers.


The challenge of determining olive oil quality is fundamental, as olive oil plays an important role in the cultural and culinary heritage of the Mediterranean countries, and its demand has grown in the latest years to other regions of the world. The growing interest, particularly in its highest quality grade, extra virgin olive oil (EVOO), is due to its high nutritional value, its richness in bioactive molecules \cite{SERRANO2021110257}, and its importance to our health due to its content of anti-inflammatory and antioxidant substances. For these reasons, extra virgin olive oil (EVOO) is a fundamental ingredient of the dietary pattern known worldwide as the "Mediterranean diet", which has been associated with important health benefits, such as the reduction of the prevalence of cardiovascular and metabolic diseases \cite{uylacser2014historical,fabiani2016anti,gorzynik2018potential}.

 
Fluorescence spectroscopy has attracted significant research efforts in the last years, as it offers a rapid, cost-efficient and at the same time sensitive technique to investigate the properties of vegetable oils \cite{karoui2011fluorescence,kongbonga2011characterization,sikorska2012analysis}.
Several fluorescent compounds are naturally present in olive oil, like pigments such as chlorophyll and beta-carotene, phenolic compounds, such as tocopherol, and primary and secondary oxidation products \cite{martin2021non}. These compounds are related to the quality criteria established in the European regulation. It is therefore of great importance to develop methods for extracting those physico-chemical information from fluorescence spectra.
The extraction of information from the spectral data can be a difficult task depending on the type of data acquired, which may range from a single spectrum to the more complex excitation emission matrices (EEMs), synchronous scanning data \cite{skoog2017principles} or near-infrared spectroscopy \cite{YUAN2020109247}.
Typical approaches consist in multivariate analysis techniques and classification methods, like for example, principal component analysis (PCA), partial least square regression (PLS), and PLS discriminant analysis (PLS-DA) to mention only a few.

The use of artificial neural networks (ANN) is known to be a useful tool, particularly because it does not require a pre-processing of the data or a dimensionality reduction \cite{Michelucci2017}. Several reviews describe the application of statistical and machine learning methods, including ANN, to the analysis and quality determination of olive oil \cite{sikorska2014vibrational,zaroual2021application,meenu2019critical,gonzalez2019critical}.
Feed-forward neural networks have been up to now successfully employed for classification purposes starting from fluorescence data  \cite{venturini2021exploration}, but do not offer sufficient flexibility for more complex tasks that analyse data that have some kind of spatial structure (like two-dimensional images or one-dimensional optical spectra).
To address this issue, various architecures, as vision transformers or convolutional neural networks, have been applied to the classification problem of vegetable oils \cite{ZHAO2022113173} with fluorescence data.

A  neural network architecture more efficient with one-dimensional input data is the one-dimensional convolutional neural network (1D-CNN) one, as recent works has shown for spectroscopic classification \cite{acquarelli2017convolutional}, electrocardiography real-time classification \cite{kiranyaz2015real}, for chemometric analysis from, for example, near-infrared reflectance spectra, and near- and mid-infrared absorption spectra \cite{malek2018one}.

By drastically reducing the requirement on the measuring hardware and on the quality of data, this work presents a novel method to extract the physico-chemical properties relevant for the quality characterization of virgin olive oil from fluorescence spectra using 1D-CNN to fluorescence spectra. The spectra need to be acquired with a very simple and compact sensor from undiluted and unprepared samples. To the best of our knowledge, this is the first time that all the key parameters are extracted simultaneously, without pre- and post-processing of the data from a simple fluorescence spectrum. The limitations and further development possibilities are also discussed in the conclusions.

The  contributions of this paper are four. 
Firstly, it describes an approach to extract the five physico-chemical characteristics relevant for the determination of olive oil's quality from one single fluorescence measurement that can be done with a low-cost sensor in less than a second. The method is described in detail with guidelines and criteria for the implementation. 
Secondly this approach does not require a technical training to use once the neural network has been trained and therefore highlight the high impact and applicability of this approach in the olive oil industry. Thirdly, by using a sensor based on low-cost components this approach highlight a highly probable democratisation of olive oil quality control.
Finally, the method is demonstrated by the application on a dataset of Spanish oils and shows, for the first time, that is possible to compete for quantitative analysis with complex chemical analysis, for example, chromatography, using a simple and fast optical measuring method supported by convolutional neural networks in one dimension.

\section{Materials and Methods}
\label{sec:material_methods}

\subsection{Olive Oil Samples}
\label{sec:OliveOilSamples}

In this study 22 virgin olive oils of three qualities were investigated: extra virgin olive oil (EVOO), virgin olive oil (VOO), and lampante olive oil (LOO). The oils were provided by the producer Conde de Benalúa, Granada, southern Spain, from the 2019-2020 harvest.
All the samples were analyzed by accredited laboratories for the chemical and organoleptic properties according to the current European regulation \cite{regulation1991commission,european2013commission}. The selected properties relevant to this study are listed in Table \ref{tab:oils}. 

\begin{table*}[hbt]
\centering
\begin{tabular}{lcccccc} 
\specialrule{.2em}{0em}{0em} 
Label	& Acidity & Peroxide value & $K_{270}$ & $K_{232}$ & Ethyl esters & Quality \\
	&  (\%) & (mEq O$_2$/kg) &  &  & (mg/Kg) &  \\
\specialrule{.1em}{0em}{0em}  
         D03 & 0.35 & 8.4 & 0.123 & 1.435 & 26 & VOO\\
         D04 & 0.34 & 8.6 & 0.108 & 1.403 & 40 & VOO\\
         D05 & 0.36 & 10.3 & 0.112 & 1.44 & 18 & VOO\\    
         D06 & 0.31 & 9.2 & 0.151 & 1.484 & 18 & VOO\\    
         D07 & 0.50 & 8.9 & 0.150 & 1.537 & 47 & VOO\\
         D08 & 0.40 & 8.5 & 0.158 & 1.546 & 25 & VOO\\
         D19 & 0.25 & 4.9 & 0.13 & 1.540 & 10 & EVOO\\
         D20 & 0.26 & 4.6 & 0.14 & 1.540 & 10 & EVOO\\
         D35 & 0.17 & 6.4 & 0.12 & 1.63 & 8 & EVOO\\
         D38 & 0.16 & 6.4 & 0.12 & 1.63 & 9 & EVOO\\
         D45 & 0.17 & 4.9 & 0.12 & 1.63 & 7 & EVOO\\
         D46 & 0.18 & 5.0 & 0.13 & 1.63 & 8 & EVOO\\
         D47 & 0.18 & 5.2 & 0.13 & 1.64 & 16 & EVOO\\
         D49 & 0.9 & 9.9 & - & - & - & LOO\\
         D51 & 2.16 & - & - & - & - & LOO\\
         D52 & 1.78 & 22 & - & - & - & LOO\\
         D53 & 0.7 & 8.7 & - & - & - & LOO\\
         D64 & 0.2 & 7.1 & 0.13 & 1.63 & 29 & VOO\\
         D73 & 0.2 & 8.9 & 0.14 & 1.66 & 15 & EVOO\\
         D77 & 0.24 & 10.4 & 0.13 & 1.74 & 26 & VOO\\
         D81 & 0.16 & 4.9 & 0.12 & 1.63 & 9 & EVOO\\
         D92 & 0.18 & 5 & 0.17 & 1.91 & 15 & EVOO\\
\specialrule{.2em}{0em}{0em} 
\end{tabular}
\caption{List of the olive oils samples analyzed in this study including selected physico-chemical characteristics. EVOO: extra virgin olive oil, VOO: virgin olive oil, LOO: lampante olive oil.\label{tab:oils}}
\end{table*}

\begin{figure*}[b!]
    \begin{center}
        \includegraphics[width=15cm]{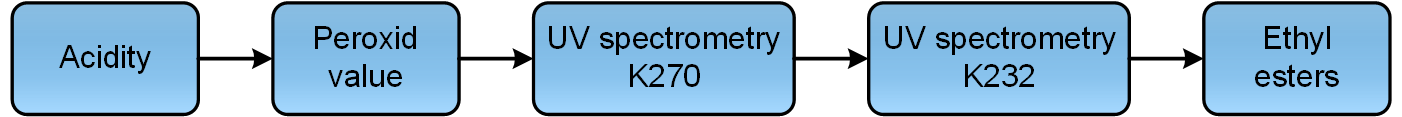}
        \caption{Sequence of parameters to be analysed for the verification of olive oil quality. 
        Adapted from \cite{regulation1991commission,european2013commission}.}
        \label{fig:decision_tree}
    \end{center}
\end{figure*}

Following the European regulation \cite{regulation1991commission} and its amendments \cite{european2013commission} the parameters used to determine the quality of the olive oil are shown schematically in Fig. \ref{fig:decision_tree}. The same parameters are investigated in this study. Note the parameter $\Delta K$ is not considered in this study since the measured values were almost identical in all the samples  within the experimental error in the measurement from the accredited laboratories.

\subsection{Instrumentation}
\label{sec:Instrumentation}
 
\begin{figure}[hbt]
    \centering
    \includegraphics[width=4.5cm]{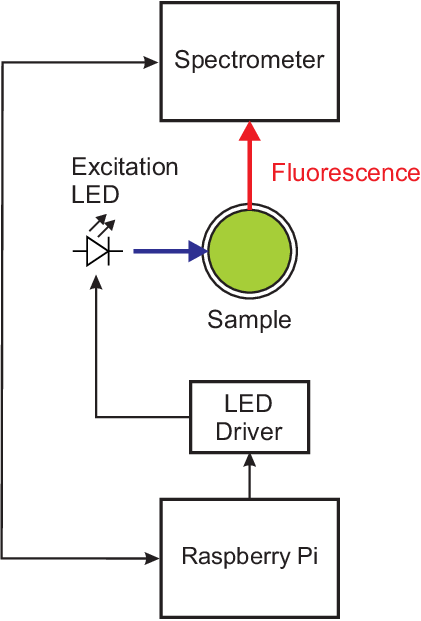}
    \caption{Schematics of the fluorescence sensor. Blue: excitation light, red: fluorescence light.}
    \label{fig:schematics}
\end{figure}

 The fluorescence spectra were taken with a sensor that has a very simple and compact design, schematically shown in Figure \ref{fig:schematics}. The excitation light is provided by an excitation UV LED. The LED can be exchanged. In this study, three wavelengths were investigated: 340 nm, 365 nm, and 395 nm. These excitation wavelengths were chosen because they correspond to an absorption maxima in the absorption band of the fluorophores present in olive oil, such as chlorophylls \cite{ferreiro2017authentication,torreblanca2019laser,borello2019determination}. The oil samples were placed into commercial transparent 4 ml glass vials, taking care that no headspace was present to reduce oxidation. 
 The fluorescence is collected by a miniature spectrometer (STS-Vis, Ocean Optics, USA) placed at 90$^\circ$ with respect to the LED to avoid the excitation light transmitted by the sample to reach the spectrometer. Both the LED driver and the spectrometer are controlled by a Raspberry Pi. The details of the device are reported in \cite{venturini2021exploration}.

All the measurements in this work were performed on undiluted samples. Although fluorescence in olive oil is subjected to the inner filter effect \cite{skoog2017principles}, the problem is not relevant for the analysis discussed in this work. In fact, the fluorescence is intense enough that the strong absorption does not decrease the signal-to-noise ratio, and possible sample-dependent effects are learned and compensated by the artificial neural network model. For each olive oil sample, 20 spectra were taken, each acquired with 1 second integration time. All the spectra were acquired under identical conditions (illumination intensity, integration time, and geometry) to be able to quantitatively compare the different intensities.

\subsection{Dataset Preparation}
\label{sec:dataset}

Since the oils were measured by different laboratories and are of different qualities, the amount of data available per oil varies. For example for some LOO oils like D49 or D52, only the acidity and peroxide value were measured. If the value of the parameter is missing, such a sample was not considered for the training and test of the ANN. Therefore, the number of oils available for the estimation of the chemical parameters depends on the parameter itself. The number of samples considered for each parameter is listed in Table \ref{tab:number_samples}.

\begin{table*}[h!]
\centering
\begin{tabular}{lc} 
\specialrule{.2em}{0em}{0em} 
Parameter	& Number of samples  \\
\specialrule{.1em}{0em}{0em} 
Acidity & 22 \\
Peroxide value & 21 \\
$K_{270}$ & 18 \\
$K_{232}$ & 18 \\
Ethyl esters & 18 \\
\specialrule{.2em}{0em}{0em} 
\end{tabular}
\caption{Number of olive oils samples used for the training and test of the CNN for for each parameter.
\label{tab:number_samples}}
\end{table*}

All the spectra are normalized after the dark background is subtracted so that each of the spectra has an average of 0 and a standard deviation of 1.

\subsection{Convolutional Neural Network Model}
\label{sec:CNN}

\begin{figure*}
    \begin{center}
        \includegraphics[width=16cm]{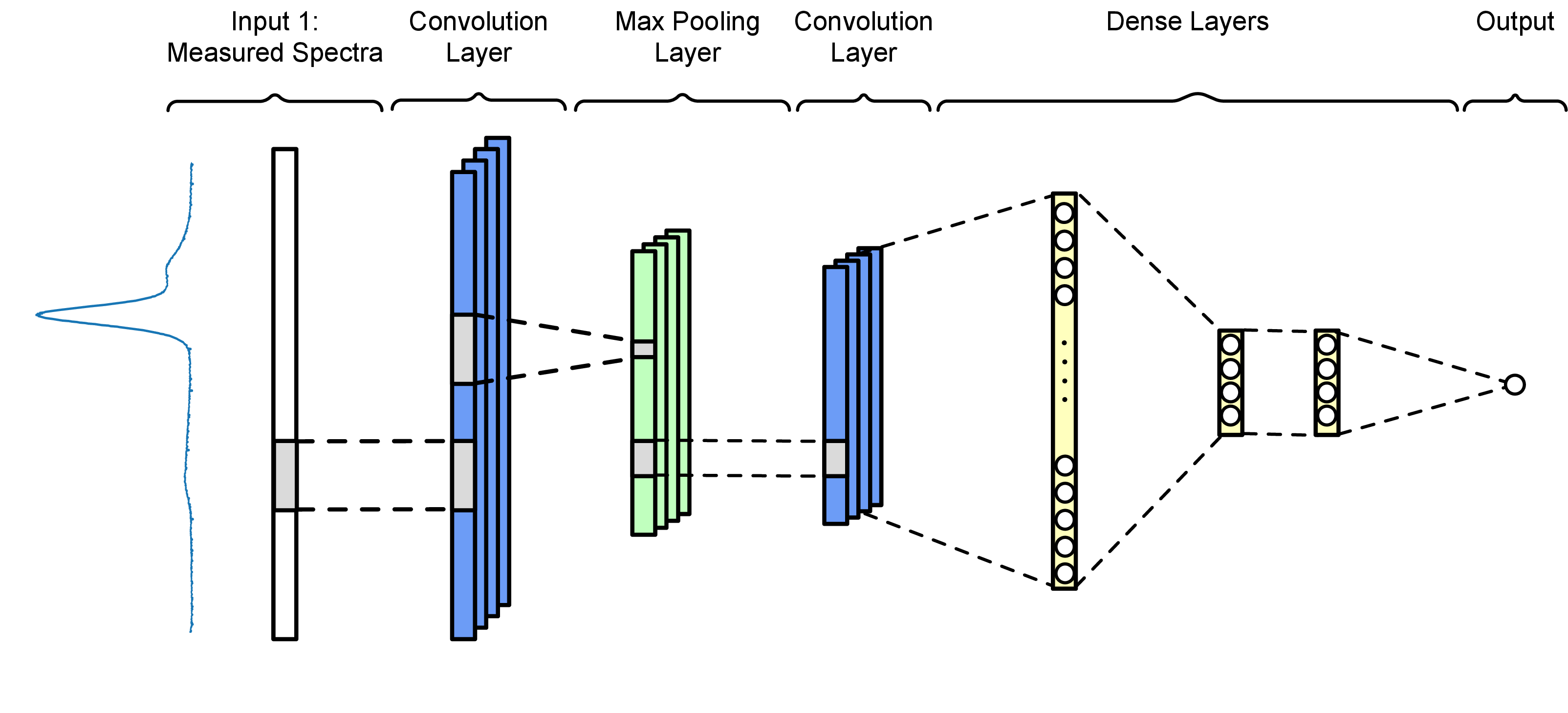}
        \caption{A schematic representation of the 1D-CNN used in this paper. The blue layers are convolutional ones, the green max pooling layers and the yellow marked ones are dense layers. The output layer has 1 neuron with the identity activation function.}
        \label{fig:1dcnn}
    \end{center}
\end{figure*}

The model developed for this work is shown in Figure \ref{fig:1dcnn} and consists of a one-dimensional convolutional neural network (1D-CNN) with one convolutional layer, followed by a max-pooling and a second convolutional layer with finally two dense layers and an output layer with one single neuron with the identity activation function. This choice was inspired by previous studies, where 1D-CNNs with two or three convolutional layers were applied to different spectroscopic data, as for example reflectance spectra and Raman spectra \cite{malek2018one,zhang2019deepspectra,liu2018transfer}.
The idea behind the sequence of layers is that the first layer extracts rough data patterns, and the subsequent layers learn more high-level abstractions. A convolutional layer is characterized by the number of filters and their size. During the 1-D convolution operation, each filter is convolved across the length of the input array, computing the dot product between the filter entries and the input, producing a 1-dimensional array (called feature map) for each of the filters \cite{michelucci2019advanced}.

In a CNN the learnable parameters are the filter themselves that are learned by backpropagation \cite{michelucci2019advanced,lecun1989handwritten,gu2018recent}. 
The parameters varied and tested in this work were the number of filters in the first convolutional layer (4 and 6), the number of filters in the second convolutional layer (4 and 6), the pooling size (8 and 16), the number of epochs (5000, 10000) and the mini-batch size (8, 16 and 64).

The size of the filters and their numbers was chosen based on the spectra and system characteristics. Previous studies suggest that the number of expected features contained in the fluorescence spectra of olive oils is of the order of 4: possible examples are the height of the main fluorescence peak, its width, the area under the peak, and area under the second fluorescence peak \cite{torreblanca2019laser, el2020rapid}. For this reason, the number of filters to test was chosen to be 4 or 6. The CNN architecture that will be selected with the hyper-parameter tuning process is expected to have a number of features used for regression (the number of feature maps, or in other words the output of the second convolutional layer) consistent with the literature. Additionally, since the spectrometer resolution is of ca. 30 pixels, the size of the filters was chosen to be 40. This reflects the fact that spectral features with a bandwidth smaller than the resolution of the spectrometer are convolved with the instrument response function. Choosing a size of 40 pixels for the filters prevents the network from considering much too granular information that the spectrometer cannot extract due to its resolution, with the additional positive effect that  overfitting will be reduced. The layers are designed to perform  feature extraction, and indirectly a dimensionality reduction, so to extract a very low number of features, by doing first max pooling and then a second convolution operation with filters of half the size of the first convolution. At the end, two small dense layers have the task to perform the regression to finally extract the chemical parameter selected.

\subsection{Metrics, performance evaluation and validation}
\label{sec:metrics}

The metrics used to evaluate the model performances are two: the mean squared error (MSE) and the mean absolute error (MAE). The MSE was used as loss function for the training of the neural networks \cite{Michelucci2017}, while the MAE was used to determine the prediction performance of the neural network. Indicating the expected (true) value of the parameters for the $i^{th}$ spectrum and the predicted value from the neural network with $y_i$ and $\hat y_i$ respectively, the two metrics can be expressed with the following formulas:
\begin{equation}
    \begin{aligned}
        \textrm{MSE} & = \frac{1}{N} \sum_{i=1}^N (\hat y_i - y_i)^2 \\
        \textrm{MAE} & = \frac{1}{N} \sum_{i=1}^N |\hat y_i - y_i| \\
    \end{aligned}
\end{equation}
where $N=20 \cdot N_{oil}$ is the number of spectra composing the dataset ($N$ is the product of 20 repetitions for each of the $N_{oil}$ oils measured). Since the dataset is small, a leave-one-out cross-validation approach \cite{make3020018} was used to determine the generalisation properties of the network. In such an approach the (20) spectra of one single oil are removed from the dataset and used for validation, while the network is trained on the spectra of all remaining oils. This procedure is repeated for each oil, therefore resulting in $N$ values of the metrics evaluated all the oils. The results reported in this paper are thus the average $\langle\textrm{MAE}\rangle$ and standard deviation $\sigma(\textrm{MAE})$ of $N$ values. A risk of the leave-one-out cross-validation is that the neural network may simply learn to predict the value of the parameter corresponding to the oil left out for all the oils. Therefore, it is quite important to always check training predictions to make sure that $\langle\textrm{MAE}\rangle$ evaluated on the training and validation dataset are comparable. For each of the $N_{oil}$ in the leave-one-out cross-validation two models during training were saved: the one with the lowest value of the loss function evaluated on the validation set (the left out oil), and the one with the lowest value of the loss function on the training set (with $N_{oil}-1$). The one that showed comparable values for $\langle\textrm{MAE}\rangle$ for training and validation dataset was then chosen.


To choose which set of hyper-parameters (number and size of filters, pooling size, epochs, etc.) normally one would select the network parameters that give the lowest value of the chosen metric (in this case  $\langle\textrm{MAE}\rangle$ on the validation dataset). However, this approach cannot be used directly here, as there is some variability (measured by the variance of the MAE $\textrm{Var}(\textrm{MAE})$) within the results and many of the calculated averages overlap within one standard deviation. 
Therefore, is important to determine if the different models in the hyper-parameter-tuning phase give results that are statistically different. This can be checked with a $t$-test has described in detail in Appendx \ref{app:stat}.
The results showed that changing the number of the filters and their size gives results that are not significantly different 
, therefore by using the Occam's razor decision criteria \cite{hiroshi} the simplest network was chosen for the final runs showed in this paper. The chosen network has 6 filters of size 40 in the first convolutional layer, and 4 filters with size 20 in the second convolutional layer. A decreasing number of filters in the first and second layers (6 and then 4 respectively) was chosen to facilitate a progressive and more stable feature extraction process \cite{Michelucci2017}.
Finally, a pooling size of 8 and a dropout rate of 0.5 were taken.

10000 epochs produced better results than 5000 consistently, therefore the former value was chosen. The mini-batch size did not influenced the results in any discernible fashion for the model that had the lowest value of the loss function evaluated on the validation dataset, therefore the value of 64 was chosen in that case. For the models that had the lowest value of the loss function evaluated on the training dataset, a mini-batch of 16 was chosen, as it showed the best and most stable results.

\section{Results and discussion}

\subsection{Fluorescence spectra of olive oil}

\begin{figure*}[hbt]
    \centering
    \includegraphics[width=14cm]{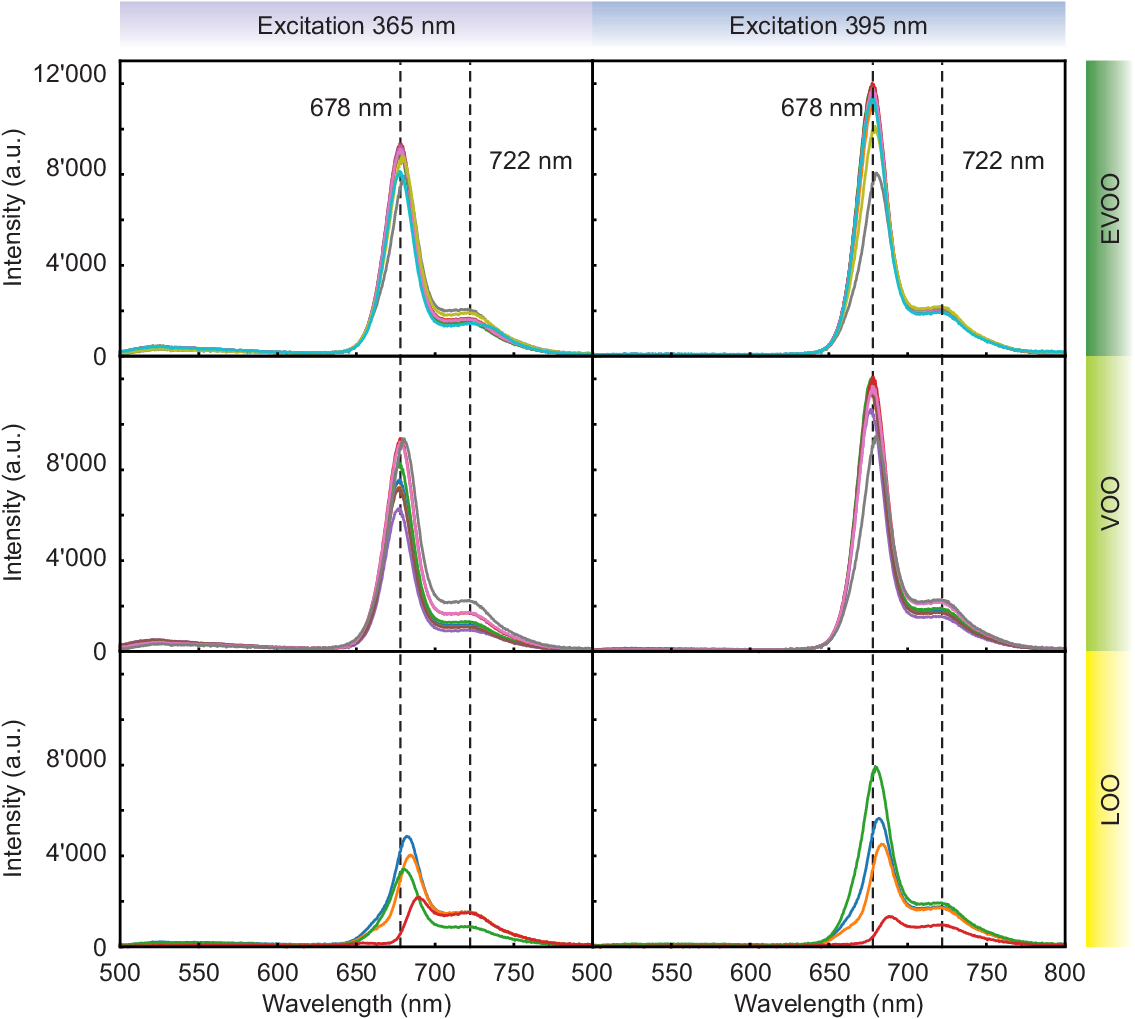}
    \caption{Fluorescence emission spectra of the measured olive oils divided in the quality classes EVOO, VOO and LOO. On the left: spectra obtained with excitation at 365 nm; on the right: spectra obtained with excitation at 395 nm. Each curve shows a single spectrum without averaging or smoothing.}
    \label{fig:spectra}
\end{figure*}

The fluorescence signals at 340 nm were very weak and are hardly detectable with the simple device used in this study. For this reason, they are not reported here.
The raw fluorescence spectra of all the oils obtained with excitation at 365 nm and at 395 nm are shown in Figure \ref{fig:spectra}. For clarity, the spectra are shown divided into the three quality classes EVOO, VOO, and LOO. Each curve of Figure \ref{fig:spectra} shows one single spectrum after background subtraction, without averaging or smoothing.

The fluorescence spectrum of all oils is characterized by a strong intensity in the region between 650 nm and 750 nm, with an intense peak at ca. 678 nm and a weaker broader one at ca. 722 nm, typical of chlorophyll and pheophytins \cite{hernandez2017fast,mishra2018monitoring,baltazar2020development,galeano2003simultaneous}. The strongest peak, however, shows variations in the spectra position and intensity towards higher wavelengths, which are particularly significant in LOOs. These variations are consistent with previous results \cite{torreblanca2019laser}. The spectra obtained with excitation at 365 nm and 395 nm are very similar, with slighter higher fluorescence intensities for 395 nm excitation. This is consistent with the stronger absorption expected around 400 nm \cite{torreblanca2019laser,borello2019determination}. Noticeably, the fluorescence intensity below 650 nm is present only in spectra obtained with excitation at 365 nm and is characterized by a weaker absorption peak at ca. 525 nm, previously attributed to vitamin E \cite{kyriakidis2000fluorescence}.

\subsection{Artificial Neural Networks Results}

\begin{figure*}[b!]
    \centering
    \includegraphics[width=14cm]{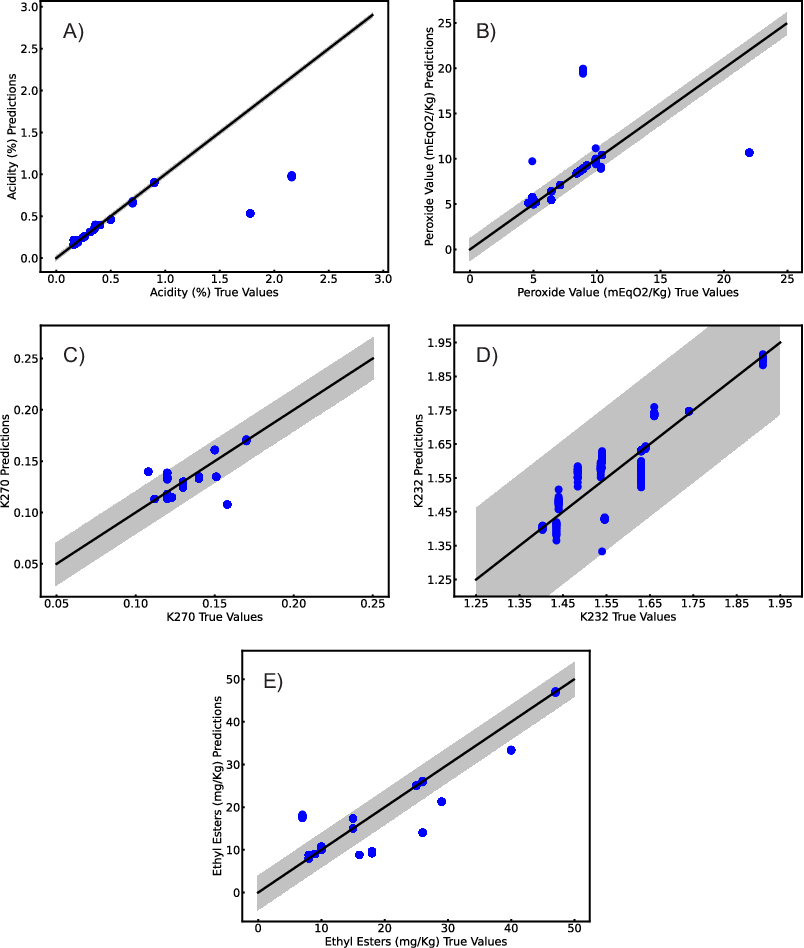}
    \caption{Comparison of the predicted and true values for all the parameters. Panel A) acidity, panel B) peroxide value, panel C) $K_{270}$, panel D) $K_{232}$ and panel E) ethyl esters. The solid line corresponds to predictions equal to the true labels. The grey area illustrates the experimental error on the true values. The yellow area marks the range of acceptability for EVOO.}
    \label{fig:predictions}
\end{figure*}

To analyze the performance of the 1D-CNN, the predicted values of the parameters were first plotted against the true values. The results are illustrated in Fig. \ref{fig:predictions}. The grey area in each panel marks the uncertainty on the true values due to the experimental error, calculated as average of the error reported by the accredited laboratory on the measured value. The yellow area  marks the range of acceptability for EVOO.

Fig. \ref{fig:predictions} panel A) shows that the 1D-CNN can predict the acidity exceptionally well, with the exception of two LOO, D51 and D52, which have values well above the 0.8\% limit for EVOO. This can be easily understood due to the lack of samples from which the ANN can learn for acidity values above 1\%: since the cross-validation is performed with a leave-one-out method, the ANN has only one single oil to learn from for acidity values above  1\%.

Fig. \ref{fig:predictions} panel B) shows the results for the prediction of the peroxide value. Also in this case the 1D-CNN can predict the value of the parameter exceptionally well. With exception of the LOO D52 and two other oils, all the predictions are within the average measurement error.

In panels C) an D) of Fig. \ref{fig:predictions}  the predictions for the two UV-spectroscopy parameters $K_{270}$ and $K_{232}$ are shown. For these two parameters, the experimental error is much larger, which means the labels used in the training phase are affected by an error. The predictions nevertheless remain very well within the grey area showing that the 1D-CNN can learn also in these cases to predict both UV-spectroscopy parameters within experimental error.

Finally, panel E) shows the performance for the prediction of the ethyl esters. Here the 1D-CNN correctly predicts several oils but has more difficulties in the prediction of others. The authors attribute part of the problem to the limited number of oils, but also to the uncertainty of the labels. Differently from the other parameters, the ethyl esters measured by the accredited laboratories were reported with errors ranging from $\pm$2 to $\pm$8 mg/kg, and in some cases without error. Also, for the ANN to learn from the spectra, the parameter must possess a direct or indirect physico-chemical signature in the fluorescence. Due to the simplicity of the sensor of this study, the fluorescence signature may be insufficiently strong or clear. Nevertheless, the method described here can give a fast and inexpensive qualitative indication of the ethyl esters without the use of gas chromatography.

The analysis at 365 nm is very similar to the one performed at 395 nm, suggesting that similar information is contained in the spectra. The use of both 365 nm and 395 nm spectra was found to be more prone to overfitting without improving the prediction performance.

The results can be quantified by calculating the metric $\langle\textrm{MAE}\rangle$ and its standard deviation $\sigma(\textrm{MAE})$, evaluated with leave-one-out cross-validation on both the training and the validation dataset. The results for all the parameters can be found in Table \ref{tab:NN_results}. In the table are also reported the average error, calculated as the MAE divided by the true label for every single oil and then averaged for all the oils, and the label error, calculated as the experimental error divided by the true label for every single oil and then averaged for all the oils.
  
\begin{table*}[h!]
\centering
\begin{tabular}{lcccc||cc}
\specialrule{.2em}{0em}{0em} 
   Parameters & $\langle\textrm{MAE}_T\rangle$ & $\sigma(\textrm{MAE}_T)$ &  $\langle\textrm{MAE}_V\rangle$  & $\sigma(\textrm{MAE}_V)$ & 
   \begin{tabular}{@{}c@{}}Average\\ error (\%)\end{tabular}
   &\begin{tabular}{@{}c@{}}Label\\ error (\%)\end{tabular}\\
\specialrule{.1em}{0em}{0em} 
Acidity (\%)& 0.10  & 0.05 & 0.12 & 0.35 & 10 & 8 \\
Peroxide Value (mEqO2/Kg)  & 1.01 & 0.65 & 1.31 & 3.19 & 12 & 17 \\
$K_{270}$ & 0.008 & 0.003 & 0.010 & 0.013 & 7 & 15 \\
$K_{232}$ & 0.03 & 0.02 & 0.04 & 0.04 & 2.5 & 13 \\
Ethyl Esters (mg/Kg) & 3.1 & 1.6 & 3.6 & 4.3 & 23 & 28 \\
\specialrule{.2em}{0em}{0em} 
\end{tabular}
\caption{Performance comparison the different neural network architectures. $a$ indicates the accuracy, $\langle a\rangle$ its mean, and $\sigma(a)$ its standard deviation evaluated over 10 different splits. T: Training, V: Validation.}
\label{tab:NN_results}
\end{table*}

Comparing the $\langle\textrm{MAE}\rangle$ for the training and for the validation shows that the chosen models were robust and did not incur in the risks associated with the leave-one-out cross-validation described. Additionally, Table \ref{tab:NN_results} shows that the average error in all cases is lower (for the parameters peroxide value and ethyl esters) or much lower (for the parameters $K_{270}$ and $K_{232}$) than the experimental error from the measurements of the accredited laboratories. Only for the acidity, the average error in the prediction is slightly higher then the label error.

The results illustrated in Table \ref{tab:NN_results} demonstrate that from a simple fluorescence spectrum, acquired with a simple and compact sensor, it is possible to predict within the typical experimental error all the chemical parameters relevant for the quality assessment of olive oil. Note that the experimental error that has been used in Table \ref{tab:NN_results} is the average of the errors provided by the accredited laboratory, thus larger errors occurs quite frequently. The values in Table \ref{tab:NN_results} for the label error column should therefore be considered an optimistic evaluation of the typical error.
The limitations to the performance observed in this study are due to the limited number of oils available for the training and the distributions of the values of the parameters (most clearly seen for the acidity, where only two oils have values in upper range).
On the other hand, it must be noted that the single origin of the olive oil samples, and thus their similar chemical characteristics, makes the task of extraction of chemical parameters somewhat easier. For a more heterogeneous dataset of olive oils it is expected that a more complex architecture will be necessary, as well as a larger dataset.

\section{Conclusions}

The results in this paper show clearly how the proposed method could substitute a more complex chemical analysis for regular quality assessment and help olive oil producers in keeping the quality of their oils under continuous control. The 1D-CNN used in this work was designed to account for the sensor characteristics (e.g., resolution) and the knowledge of the problem (e.g., expected number of features in the spectrum). As a result, the method has shown a very promising performance: from the simple fluorescence spectra it is possible to predict, within the typical experimental errors, all the five physico-chemical characteristics necessary for quality assessment of olive oil.
Of course one should note that the dataset size in this study is small and, therefore, the results should only be considered as an indication of the potential of the method. Naturally, a larger dataset would allow a more complete analysis of the generalisation properties of such models when applied to olive oils optical spectra. Nonetheless this method is extremely cheap, fast and can be done by the producers themselves on-site practically without any scientific training, except knowing how to operate a computer and put oil in a vial. 

The future potential of this approach is very exciting. For example by having multiple samples from multiple years, and using meteorological information about the geographic location of production one could correlate quality with information as amount of precipitations, temperature and so on. This would open the possibility of predicting quality based on external factors, probably one of the greatest challenge in the olive oil economy.

As briefly mentioned one of the challenge to be solved in the future is the application of this approach to olive oil samples coming from different producers, different geographical locations or from harvests of different years. It is to be expected that the chemical signatures in the phosphorescence spectra will not be similar anymore within those subgroups, making the prediction of the parameters a much greater challenge. In this case, more complex 1D-CNN architectures and larger datasets will be necessary to keep into account the heterogeneities in the olive oil samples and solve this task with the same degree of performance. Preliminary results by the authors for 1D-CNN with multiple input branches (by using producer information for example) indicates great potential in addressing this challenge. Such architectures should be able to adapt to this data complexity and should deliver a similar performance, if enough data is of course available.

Finally, this approach is of course not limited to olive oils but can be extended to other substances, making the results described here a very promising indication of what could be achieved through one-dimensional convolutional neural networks applied to optical spectra.

\section{Funding}
This work was supported by the projects: “VIRTUOUS” funded by the 
European Union’s Horizon 2020 Project H2020-MSCA-RISE-2019 Grant No. 872181;
"SUSTAINABLE" funded by the European Union’s Horizon 2020 Project H2020-MSCA-RISE-2020 Grant No. 101007702; “Project of Excellence” from Junta de Andalucia-FEDER-Fondo de Desarrollo Europeo 2018. Ref. P18-H0-4700.

\appendix
\section{Statistical testing of equivalence of averages}
Given two sets of hyper-parameters, indicated here with the subscripts 1 and 2, one can test the equality of the two means of the MAE, $\langle\textrm{MAE}_1\rangle$ and $\langle\textrm{MAE}_2\rangle$ respectively, by using the $t$-statistic \cite{hogg1977probability}. The formulas used in this paper are based on the ones for confidence intervals for the difference of the means when the variances are unknown and the sample size is relatively small. Note that the $t$-statistics technically works when one deals with normal distributions. In general, the MAE values from the leave-one-out cross-validation approaches have an unknown distribution. However, since one is considering the average, thanks to the central limit theorem, one can assume that the distribution of $\langle\textrm{MAE}\rangle$ is approximated by a normal distribution (at least one that is not too skewed) and therefore the choice of this approach is justified \cite{hogg1977probability}. $N_{oil}$ is of the order of 20 (see Table \ref{tab:oils}), a number typically considered not large enough for the central limit theorem. Nevertheless, being close to the suggested value of 30, it should give a useful estimate of the statistical significance of the average difference between different sets of hyperparameters. The null-hypothesis $H_0$ that the two means are equal is rejected if the observed value of
\begin{equation}
    T=\frac{\langle\textrm{MAE}_1\rangle-\langle\textrm{MAE}_2\rangle}{S_P\sqrt{2/N_{oil}}}
\end{equation}
where
\begin{equation}
\begin{aligned}
    S_P=&\left[\frac{(N_{oil}-1)\textrm{Var}(\textrm{MAE}_1)}{2N_{oil}-2} + \right. \\
    +&\left.  \frac{(N_{oil}-1)\textrm{Var}(\textrm{MAE}_2)}{2N_{oil}-2}  \right]^{1/2}
\end{aligned}
\end{equation}
is larger than $t_\alpha(2N_{oil}-2)$ \cite{hogg1977probability} (right-trail probability of size $\alpha$ for the $t$-distribution with $2N_{oil}-2$ degrees of freedom, or in other words the value that satisfy that the probability $P(t\geq t_\alpha) = \alpha$) for some chosen value of $\alpha$. For this work $\alpha = 0.05$ was chosen.

\bibliography{mybibfile}

\end{document}